# Enhanced Anime Image Generation Using USE-CMHSA-GAN

J. Lu

Department of Electrical & Computer Engineering, University of Washington

Seattle, WA, USA

j.lu20000701@gmail.com

## Abstract

With the growing popularity of ACG (Anime, Comics, and Games) culture, generating high-quality anime character images has become an important research topic. This paper introduces a novel Generative Adversarial Network model, USE-CMHSA-GAN, designed to produce high-quality anime character images. The model builds upon the traditional DCGAN framework, incorporating USE and CMHSA modules to enhance feature extraction capabilities for anime character images. Experiments were conducted on the anime-face-dataset, and the results demonstrate that USE-CMHSA-GAN outperforms other benchmark models, including DCGAN, VAE-GAN, and WGAN, in terms of FID and IS scores, indicating superior image quality. These findings suggest that USE-CMHSA-GAN is highly effective for anime character image generation and provides new insights for further improving the quality of generative models.

## 1. Introduction

Two-dimensional culture is gradually moving from subculture to mainstream, and comics, as an important carrier of secondary culture, are increasingly sought after by young groups. With the growing demand for personalisation and uniqueness, many young people prefer to use manga characters as their avatars on social platforms. However, these characters are usually protected by copyright, limiting their free use. At the same time, young people's desire to express their individuality through one-of-a-kind anime characters has led to a rapid increase in demand for original anime characters [1]. To meet this demand, it has become an inevitable trend to generate personalised anime characters using Generative Adversarial Network (GAN).

Although GAN has achieved remarkable results in the field of image generation, the research on anime character image generation is still scarce and the generation effect needs to be improved. In this study, a novel generative adversarial network model USE-CMHSA-GAN is proposed for anime character image generation, which introduces Upsampling Squeeze-and-Excitation (USE) module and Convolution-based Multi-Head Self-Attention (CMHSA) module in the Deep Convolutional Generative Adversarial Network (DCGAN) generator to generate more realistic images of anime characters.

The main contributions of this paper are as follows:
- A USE module and a CMHSA module were incorporated into the DCGAN architecture to improve the overall quality of generated anime character images. The effectiveness of these two modules was demonstrated through extensive testing.
- The USE module enhances channel-level attention, effectively capturing key features and outputting refined feature maps by expanding feature separation.
- The CMHSA module enables the model to integrate diverse features from multiple perspectives, improving its representational capacity and the ability to capture long-range dependencies.
- Comprehensive experiments were conducted on the anime-face-dataset. The results indicate that the modified model, combining USE and CMHSA modules, surpasses DCGAN and other models in overall quality for generating anime character images.

The paper is organised as follows: section 2 reviews the relevant research progress in this field; section 3 details the architectural design of the proposed model; section 4 shows the experimental results and analyses them; section 5 concludes the whole paper and proposes future research directions.

## 2. Related Work

Goodfellow et al. proposed the GAN model[2] to generate high-quality images through the competition between generators and discriminators, pioneering a classical generative model. However, GAN is prone to instability during the training process and leads to the phenomenon of mode collapse, while the quality of the generated images still has room for improvement. To improve this problem, Larsen et al. combined Variational Autoencoder (VAE) and GAN, and proposed the Variational Autoencoder - Generative Adversarial Network (VAE-GAN) model[3] to better generate high-quality images. Subsequently, Radford et al. proposed DCGAN[4], which introduced Convolutional Neural Network (CNN) into the generator and discriminator structure of GAN to significantly improve the quality and stability of the generated images.Arjovsky et al. proposed Wasserstein Generative Adversarial Network (WGAN)[5], which effectively mitigates the mode collapse problem by using the Wasserstein distance to replace the JS dispersion of traditional GANs.Zhang et al. introduced Self-Attention Mechanism (SAM) into the generator and the discriminator, which created a new approach to applying SAM in GANs[6]. Brock et al. proposed Big Generative Adversarial Network(BigGAN)[7] to further enhance the generative capability of GAN, which is suitable for large-scale and high-complexity generative tasks.

However, in the field of anime character image generation, most current studies still use DCGAN to generate anime character images, resulting in relatively poor generation results. For this reason, this paper proposes the USE-CMHSA-GAN model, which aims to generate high-quality anime character images, thereby enhancing the realism of the generated images and effectively solving the problem of the poor quality of existing anime character image generation.

## 3. Proposed Method

### 3.1. Introduction to the Method

The USE-CMHSA-GAN enhances the overall quality of generated anime character images by incorporating the USE and CMHSA modules into the generator of DCGAN.

**3.2. USE Module**

The USE module primarily consists of four steps: Squeeze, Excitation, Channel Weighting, and Upsampling[8].

First, Global Average Pooling is applied to aggregate information across each channel, allowing the capture of global information for every channel. Given an input feature map $X \in R^{C \times H \times W}$, the average value for each channel is computed as shown in Equation (3.1):

$$z_c = \frac{1}{H \times W} \sum_{i=1}^{H} \sum_{j=1}^{W} x_{c,i,j} \quad (3.1)$$

Here, $z_c$ represents the global average for the $c$-th channel of the feature map, resulting in the vector $z \in R^{C \times 1 \times 1}$.

Then, $z$ undergoes a series of nonlinear transformations using $\text{Conv}_1$ and $\text{Conv}_2$, followed by ReLU and Sigmoid activations, to produce the channel attention weights $A$, as shown in Equation (3.2):

$$A = \sigma\left(\text{Conv}_2\left(\text{ReLU}(\text{Conv}_1(z))\right)\right) \quad (3.2)$$

Here, $\text{Conv}_1$ reduces the number of channels from $C$ to $\frac{C}{2}$, while $\text{Conv}_2$ restores it from $\frac{C}{2}$ back to $C$. Finally, the Sigmoid activation is applied to obtain the attention weights $A \in R^{C \times 1 \times 1}$ for each channel.

In the next step, the attention weights $A$ are applied to the input features $X$, resulting in the recalibrated features $Y$ as shown in Equation (3.3):

$$Y = X \odot A \quad (3.3)$$

where $\odot$ denotes element-wise multiplication across channels.

Finally, transposed convolution is used to perform upsampling, mimicking the inverse operation of the convolutional layer. Figure 1 presents the USE module's architecture.

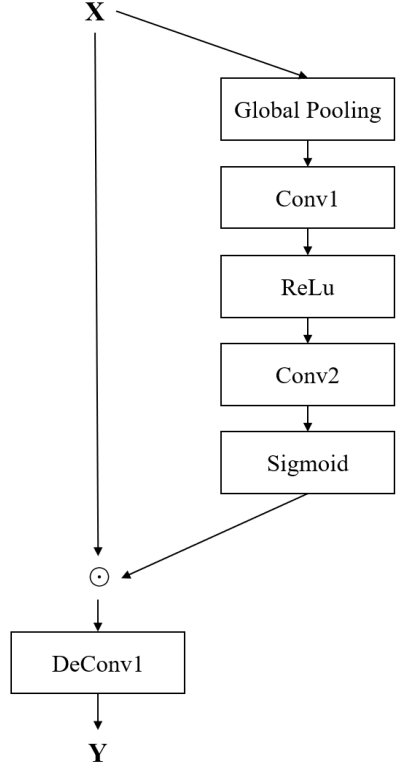

Figure 1: Architecture of the USE module

## 3.3. CMHSA Module

The core concept of the CMHSA Module is Multi-Head Self-Attention[9]. Its primary steps involve computing the similarity between the query vector $Q$ at each position and the key vectors $K$ at all other positions. This similarity is then used to apply weighted aggregation to the value vectors $A$.

Equations (3.4) and (3.5) provide the calculation formulas for the parameters $head\_dim$ and $scale$.

$$head\_dim = \frac{in\_channels}{num\_heads} \quad (3.4)$$

$$scale = \frac{1}{\sqrt{head\_dim}} \quad (3.5)$$

Equation (3.6) defines the generation of $Query$, $Key$ and $Value$:

$$Q = W_q X, \quad K = W_k X, \quad V = W_v X \quad (3.6)$$

where $X$ represents the input feature map, and $W_q$, $W_k$, $W_v$ are the weight matrices for generating $Query$, $Key$ and $Value$, respectively. These matrices are followed by a reshape operation to facilitate the computation of attention scores.

Equation (3.7) calculates the similarity between $Query$ and $Key$:

$$\text{attn}_{ij} = \frac{Q_i \cdot K_j}{\sqrt{head\_dim}} \quad (3.7)$$

Where $Q_i$ and $K_j$ represent the *Query* and *Key* vectors at different positions. A dot product operation is performed on them, followed by scaling with the *scale* parameter to control the range of values.

Then, as shown in Equation (3.8), the softmax function is applied to normalize the attention scores $\alpha_{ij}$, which represent the attention weight of position $j$ with respect to position $i$:

$$\alpha_{ij} = \frac{\exp(\text{attn}_{ij})}{\sum_k \exp(\text{attn}_{ik})} \quad (3.8)$$

The exponential function $\exp(\text{attn}_{ij})$ converts each attention score to a positive value, facilitating probability computation. The denominator $\sum_k \exp(\text{attn}_{ik})$ is the sum of the exponentiated similarity scores for all key vectors, ensuring that the total attention weight sums to 1, thus achieving normalization.

Next, dropout is applied to $\alpha_{ij}$, as shown in Equation (3.9):

$$\alpha'_{ij} = \frac{\alpha_{ij} \times \text{Mask}_{ij}}{1-p} \quad (3.9)$$

Here, $\text{Mask}_{ij}$ is a random binary mask matrix with values of 1 or 0, indicating whether each element is retained. ppp represents the dropout rate. The adjusted attention weight $\alpha'_{ij}$ introduces stochasticity to the original $\alpha_{ij}$ through dropout, helping to prevent overfitting by adding randomness.

For each position $i$, the output $Output_i$ is computed by performing a weighted sum over all input positions $j$, where the weights are given by the attention scores $\alpha'_{ij}$, as shown in Equation (3.10):

$$\text{Output}_i = \sum_j \alpha'_{ij} V_j \quad (3.10)$$

After calculating the weighted outputs for all positions $i$, they are aggregated to form the complete output tensor $Output$, which is subsequently reshaped as required.

A $1 \times 1$ convolutional layer, denoted as $out\_proj$, is used to perform a linear transformation. Assuming $out\_proj$ as the weight matrix of $out\_proj$, the final output is obtained by adding the input $X$ as a residual connection, as shown in Equation (3.11):

$$Y = W * \text{Output} + X \quad (3.11)$$

This residual connection helps mitigate the vanishing gradient problem and enhances training performance.

### 3.4. Loss Functions

The loss functions for the discriminator and generator in USE-CMHSA-GAN are designed to optimize their respective objectives, using Binary Cross-Entropy Loss.

The discriminator aims to accurately classify real and fake images, with its loss function defined in Equation (3.12):

$$L_D = -E_{x \sim p_{\text{dt}}}[\log D(x)] - E_{z \sim p_z}\left[\log\left(1 - D(G(z))\right)\right] \quad (3.12)$$

The objective of the generator is to produce fake images that resemble real ones closely enough to be identified as real by the discriminator[10]. Consequently, the generator's goal is for the discriminator's output on generated images to approach 1. Its loss function is given by Equation (3.13):

$$L_G = -E_{z \sim p_z}[\log D(G(z))] \quad (3.13)$$

**3.5. Architecture of USE-CMHSA-GAN**

The overall architecture of USE-CMHSA-GAN is consistent with that of a conventional GAN, comprising two main components: a generator and a discriminator. Through adversarial training between these components, the generator is driven to produce increasingly realistic images. The detailed structure is illustrated in Figure 2.

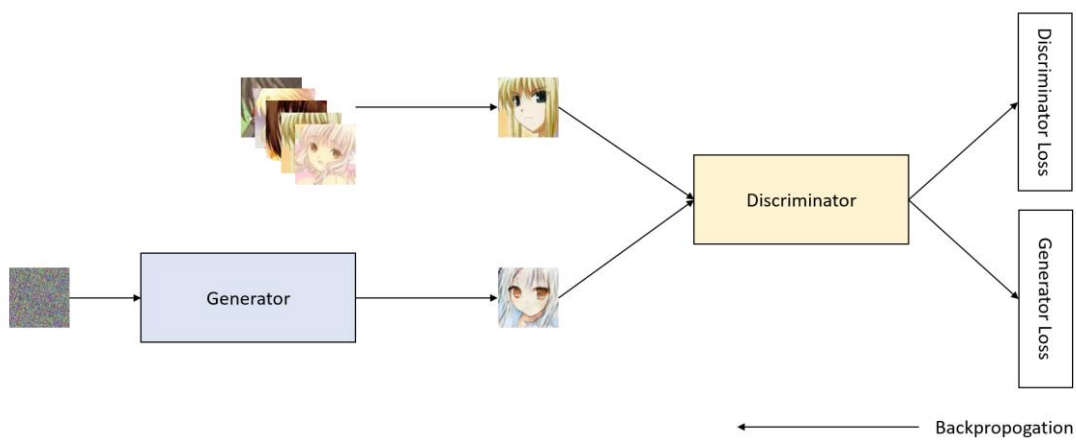

Figure 2: Overall Architecture of USE-CMHSA-GAN

In terms of implementation, USE-CMHSA-GAN is an improved version of DCGAN, with the generator enhanced by incorporating the USE and CMHSA modules. The addition of these two modules enhances feature extraction and generation quality, ultimately improving the realism of generated images. The detailed structure is shown in Figure 3.

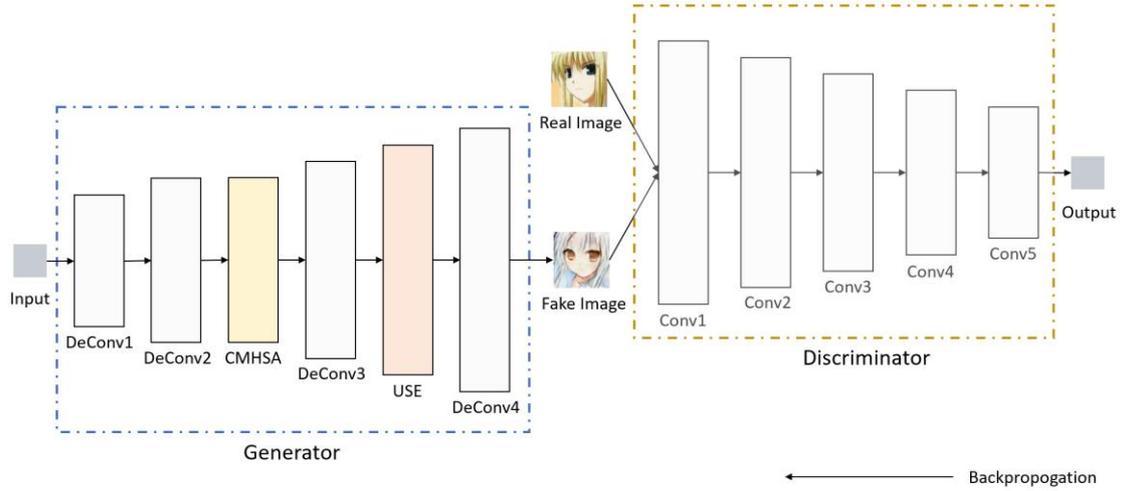

Figure 3: Detailed Architecture of USE-CMHSA-GAN

The generator in DCGAN primarily consists of Deconvolutional Layers (DeConv). In USE-CMHSA-GAN, we replaced one of the DeConv layers with the USE module and inserted the CMHSA module between two DeConv layers. This modification enhances the generator's ability to capture image features, thereby improving the realism of generated anime character images. On the discriminator side, USE-CMHSA-GAN remains consistent with DCGAN without any further modifications.

## 4. Experiment

The dataset used in this experiment is anime-face-dataset[11], which contains a total of 27,588 anime character images with a resolution of 256 × 256. We filtered the dataset to remove some lower quality images to ensure the training effect. An 80:10:10 split was applied to divide the dataset into training, validation, and test sets[12]. Before training the model, we resized the images to 64×64 and applied standard preprocessing steps, including tensor conversion and normalization, to meet the input requirements of the model and improve training efficiency.

The quality of anime character images generated by the model is evaluated using two metrics: Inception Score (IS)[13] and Fréchet Inception Distance (FID)[14]. IS measures the realism and diversity of generated images[15] by calculating the entropy of the class distribution and the conditional entropy given the image. A higher IS indicates that the generated images are more distinct in terms of category and exhibit a broader distribution across different classes. FID assesses image quality by quantifying the distance between the feature distributions of generated images and real images[16]. A lower FID suggests that the distribution of generated images is closer to that of real images, indicating higher fidelity.

### 4.1. Generated Samples

As shown in Figure 1, the left side is the real anime character image and the right side is the virtual anime character image generated by USE-CMHSA-GAN. It can be seen that USE-CMHSA-GAN has been able to accurately depict the eyes and hair part of the character during the generation process. However, the details of eyebrows, nose and mouth are shown relatively less, which is closely related to the image characteristics of the dataset. In the dataset, the focus of the images tends to be concentrated on the eyes and hair parts, while the depiction of the eyebrows is usually not obvious enough; meanwhile, some of the images have only a simple outline of the nose and mouth, which are not shown as the focus.

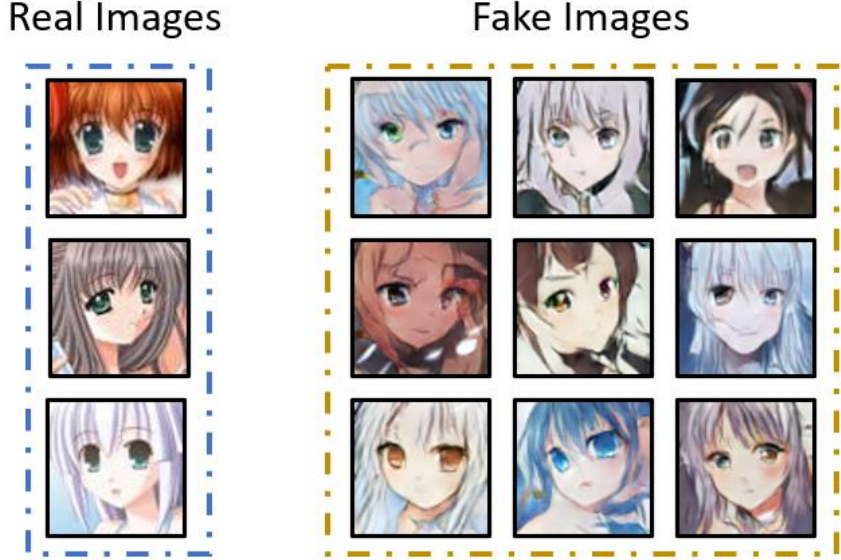

Figure 4: Real Images and Fake Images generated by USE-CMHSA-GAN

## 4.2. Model Comparison

We compared USE-CMHSA-GAN with two other models: VAE-GAN and WGAN. The anime-face-dataset was employed for training, validation, and testing of all three models. The quality of generated images was measured by FID and IS, with results shown in Table 1[17]. As shown, USE-CMHSA-GAN achieved the lowest FID and the highest IS among all models, indicating that it outperformed the others in terms of image quality. However, there is still room for improvement in the generated image quality across all models. This limitation may stem from the relatively small and inconsistent quality of the dataset samples.

| Model | FID | IS |
|---|---|---|
| VAE-GAN | 64.45 | 2.60 |
| WGAN | 79.34 | 2.35 |
| USE-CMHSA-GAN | **53.74** | **2.85** |

Table 1: Performance Comparison of Different Models on Anime-Face-Dataset

## 4.3. Ablation Study

USE-CMHSA-GAN is obtained by improving on DCGAN, and the USE module and CMHSA module are introduced in the generator part of DCGAN. As can be seen from the following table, the introduction of both the USE module and the CMHSA module can individually improve the image generation quality of DCGAN, while the model effect is further improved by combining the two, which proves the effectiveness of the USE module and the CMHSA module. The specific metrics are detailed in Table 2.

| Model | FID | IS |
| --- | --- | --- |
| DCGAN | 63.92 | 2.52 |
| USE-GAN | 58.99 | 2.69 |
| CMHSA-GAN | 55.82 | 2.68 |
| USE-CMHSA-GAN | **53.74** | **2.85** |

Table 2: Ablation Study of USE and CMHSA Modules in USE-CMHSA-GAN

## 5. Conclusion

In this thesis, we propose the USE-CMHSA-GAN model to enhance the image generation capability by adding the USE module and CMHSA module to the generator of DCGAN to generate higher quality images of anime characters. The experiments achieve excellent performance on the anime-face-dataset dataset, and the FID scores and IS scores of the generated images outperform those of DCGAN, WGAN, and VAE-GAN models, which verifies the effectiveness of the USE module and the CMHSA module in enhancing the quality of the anime character image generation.

However, the limited number of samples in the dataset used in this study and the variable quality of the image data in the dataset still impose some limitations on the quality and diversity of the images generated by the model. Therefore, future research could explore larger and more diverse anime character image datasets to further improve the generation results. Meanwhile, future work could also focus on optimising the discriminator structure of USE-CMHSA-GAN to further improve the performance and adaptability of the model.